\documentclass[letterpaper, 10 pt, conference]{IEEEconf}
\IEEEoverridecommandlockouts
\usepackage{cite}
\usepackage{amsmath,amssymb,amsfonts}
\usepackage{algorithmic}
\usepackage{graphicx}
\usepackage{textcomp}
\usepackage{mathtools}
\usepackage{multirow}
\usepackage[table,xcdraw]{xcolor}
\usepackage{adjustbox}
\usepackage{subcaption}
\usepackage{booktabs}
\usepackage{breqn}

\def\BibTeX{{\rm B\kern-.05em{\sc i\kern-.025em b}\kern-.08em
    T\kern-.1667em\lower.7ex\hbox{E}\kern-.125emX}}
\begin{document}

\title{\LARGE \bf RGB-X Classification for Electronics Sorting\\
\thanks{$^{1}$Robotics Institute, Carnegie Mellon University, 5000 Forbes Avenue, Pittsburgh, USA {\tt\small abhiman2@andrew.cmu.edu, choset@andrew.cmu.edu, mtravers@andrew.cmu.edu} }
\thanks{$^{2}$Apple Inc.}}

\author{FNU Abhimanyu$^{1}$, Tejas Zodage$^{1}$, Umesh Thillaivasan$^{2}$, Xinyue Lai$^{1}$, Rahul Chakwate$^{1}$, \\Javier Santillan$^{2}$, Emma Oti$^{2}$, Ming Zhao$^{2}$, Ralph Boirum$^{1}$ Howie Choset$^{1}$, Matthew Travers$^{1}$}

\maketitle

\begin{abstract} 

Effectively disassembling and recovering materials from \textcolor{black}{waste electrical and electronic equipment (WEEE)} is a critical step in moving global supply chains from carbon-intensive, mined materials to recycled and renewable ones. \textcolor{black}{Conventional} recycling processes rely on shredding and sorting waste streams, but for WEEE, which is comprised of numerous dissimilar materials, we explore targeted disassembly of numerous objects for improved material recovery. \textcolor{black}{Many} WEEE objects \textcolor{black}{share many key features and therefore can} look \textcolor{black}{quite} similar, but their material composition and internal component layout can vary,  and thus it is critical to have an accurate classifier for subsequent disassembly steps for accurate material separation and recovery. This work introduces RGB-X, a multi-modal image classification approach, that utilizes key features from external RGB images with those generated from X-ray images to accurately classify electronic objects.  More specifically, this work develops Iterative Class Activation Mapping (iCAM), a novel network architecture that explicitly focuses on the finer-details in the multi-modal feature maps that are needed for accurate electronic object classification. \textcolor{black}{In order to train a classifier, electronic objects lack large and well annotated X-ray datasets} due to expense and need of expert guidance. To overcome this \textcolor{black}{issue}, we present a novel way of creating a synthetic dataset using domain randomization applied to the X-ray domain. The combined RGB-X approach gives us an accuracy of 98.6$\%$ on 10 generations of modern smartphones, which is greater than their individual accuracies of 89.1$\%$ (RGB) and 97.9$\%$ (X-ray) independently. We provide experimental results\footnote[3]{Experimental work done at Biorobotics Lab, Robotics Institute, Carnegie Mellon University} to corroborate our results. 
\end{abstract}

\vspace{-0.2cm}
\section{Introduction} \label{Introduction}

Daisy\cite{Daisy2018},\cite{Daisy2019} and Dave\cite{Dave2020},\cite{EnvironmentalReport2020}, Apple's existing disassembly robots, \textcolor{black}{disassemble} electronic devices and components to enable the recovery of \textcolor{black}{precious} materials like rare earth elements, steel, and tungsten.  Diverse object disassembly involves hundreds of decisions throughout the process, and  object classification is the first major decision in the pipeline. In practice, precisely classifying electronics objects is challenging as different objects \textcolor{black}{of similar product line} often have subtle visible differences leading to incorrect classifications that can cause sub-optimal material recovery if material compositions \textcolor{black}{differ} between the actual and predicted class. 
\textcolor{black}{Conventional} machine vision has to be highly accurate for \textcolor{black}{recycling}, however to classify all desired WEEE objects for disassembly using one line also requires that the system be capable of handling high variations seen in each class of multi-class WEEE such as cracks, bends, missing or replaced components, or other deformations that are difficult to predict and that inhibit conventional classification methods (e.g., RGB-based methods). These variations eliminate methods like weighing when numerous classes differ by only a few grams. 
Thus, this paper addresses challenges in classifying WEEE as they have subtle differences on the outside forcing a classifier to focus on extremely specific details on the outside or look inside for other distinguishing features. 
 
\begin{figure}[t]
    \begin{center}
        \includegraphics[width=0.8\linewidth]{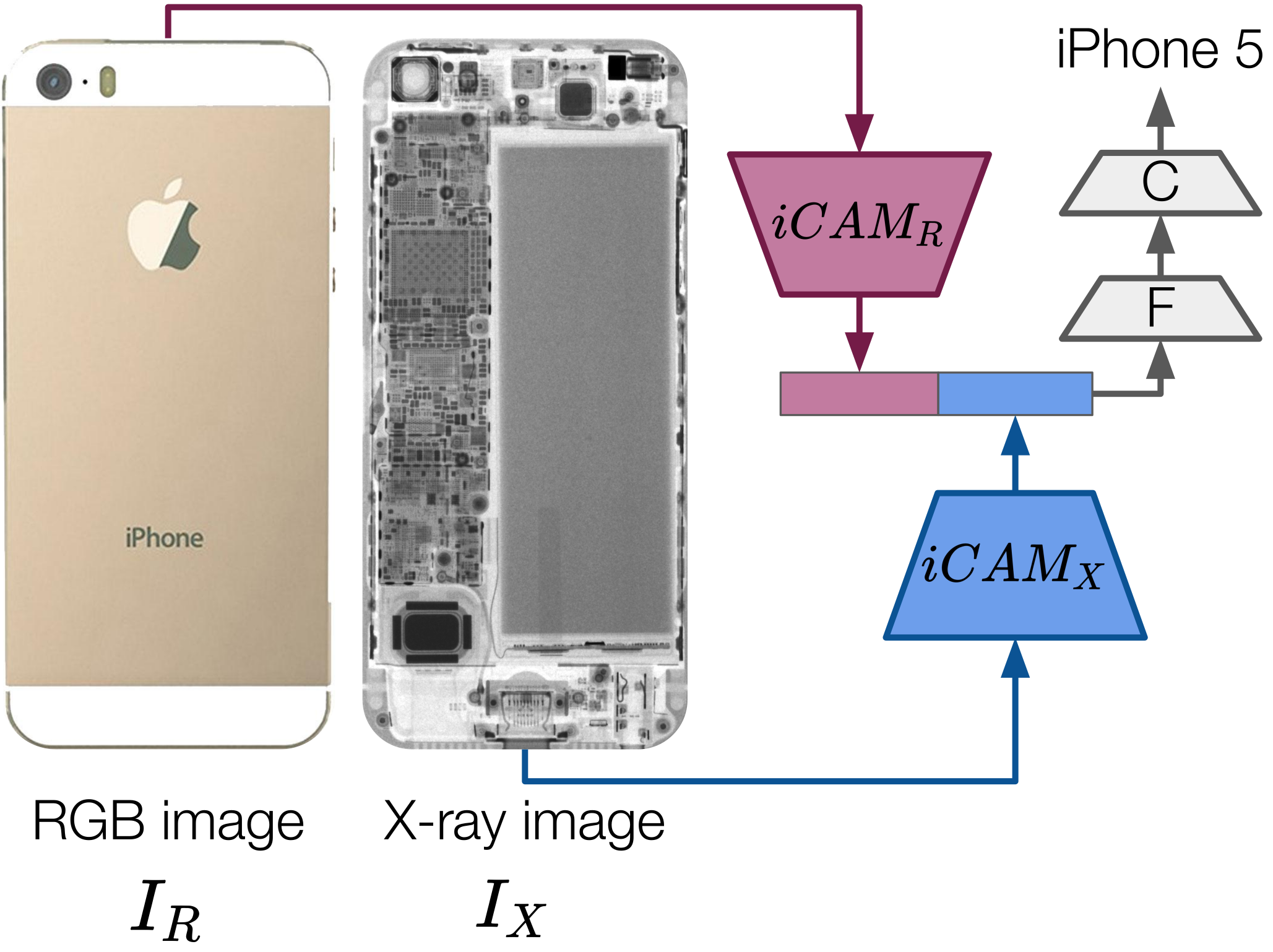}  
    \end{center}
    \caption{Our approach fuses feature maps of RGB ${I_R}$ and X-ray image ${I_X}$ in order to capture both internal and external features of electronics devices. We develop a novel architecture, $iCAM$ for suitable feature maps extraction of electronics images and show that the RGB-X fusion along with $iCAM$ gives classification accuracy as high as 98.8\%.} 
    \label{fig:pipeline_image}
     \vspace{-0.7cm}
\end{figure}

 \vspace{1cm}
\begin{figure*} 
    \begin{center}
        \includegraphics[width=0.9\linewidth]{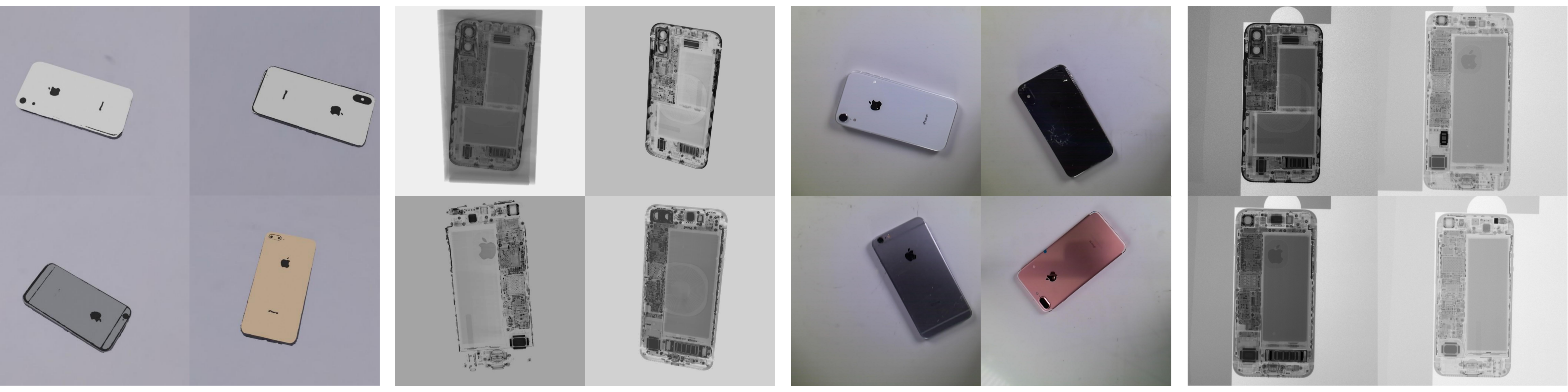}
    \end{center}
    \caption{Examples of the training and the testing data for the RGB and the X-ray images. From left to right: synthetic RGB images (training), synthetic X-ray images (training), real RGB images (testing), real X-ray images (testing). } 
    \vspace{-0.4cm}
    \label{fig:dataset_image}
\end{figure*}

\vspace{-1cm}

In this work, we present a novel attention based iterative strategy, Iterative Class Activation Mapping (iCAM), to localize and then extract key features \textcolor{black}{of relevant components} from inside and outside of the devices for classification. This effectively guides the models to focus on the appropriate features for better decision making.

Also, to leverage both outside
and inside features of electronic objects, we propose
RGB and X-ray classification modalities individually and
then uniquely combine them for their classification. Though there has been previous work on multi-domain classification, to the best of our knowledge, this work is the first to combine RGB and X-ray domains to perform classification for recycling purposes. This combination helps find common key features among the internal and external parts of the objects, making the classification robust to conditions like lighting, or wear and tear of the device.

The single-mode \textcolor{black}{(RGB mode or the X-ray mode)}, as well as the joint training phase, needs an annotated dataset for training. Unlike other areas of image classification, used electronics lack a well-annotated dataset for training an image classifier robustly, especially in the X-ray domain. The creation of annotated X-ray data is tedious and requires expert guidance. Thus, in this work, we propose a data randomization pipeline to generate synthetic X-ray data for training. We create synthetic X-ray images by projecting 3D-Computed Tomography (CT) scans into a 2D space and randomizing the relevant parameters like orientation, intensity settings, noise and background. 


The paper is organized as follows: Section \ref{prior_work} introduces the prior work reported for electronic objects' image classification. Section \ref{Approach} provides a detailed description on the dataset, iterative class activation mapping (iCAM) architecture and the training strategy. Section \ref{results} describes the experimental results on RGB-X classification. Section \ref{conclusion} discusses the conclusion and the future work.
\section{Prior Work} \label{prior_work}
This section discusses the existing WEEE classification methods. It also discusses the prior work regarding the dataset generation process, fine-grained image classification and multi-modal classification.

\subsection{Classification for WEEE sorting}\label{AA}
WEEE is an existing problem where both classical techniques as well as deep learning architectures have been used recognize objects. UNU-KEYS \cite{UNU-KEY2018} is a classification method that is being used to classify WEEE by using attributes such as average weight, material compositions and  end-of-life characteristics. This works well for coarse level classification, but does not work well for similar looking devices which have subtle feature differences. To overcome that, state of the art deep learning methods have also being used to accurately classify WEEE items. Standard network architectures like Faster Squeeze-Net \cite{Xu2020AnEC} and YOLOv3 \cite{Kumar2020} were used to sort electronics. The Swedish company, Refind, in collaboration with the Danish Institute of Technology has also demonstrated the feasibility of sorting different types of printed circuit boards, mixed electronic scrap, and batteries by capturing color images and applying a deep learning architecture \cite{DTI2017},\cite{Refind2019}. Use cases include applying deep learning to sort different types of coin cells \cite{Karbasi2018}, and additional imaging inputs such as thermal and X-ray imaging are being explored for WEEE recognition \cite{Gund2018}, \cite{SterkensWouter2020Daro}. While the prevalence of deep learning has allowed for advances in classification, these methods are still primarily used for component level classification and they use a single mode for classification, therefore making models sensitive to real world scenarios like lighting, noise, and deformations. 
\begin{figure*}
    \begin{center}
        \includegraphics[width=0.9\textwidth, scale=0.8]{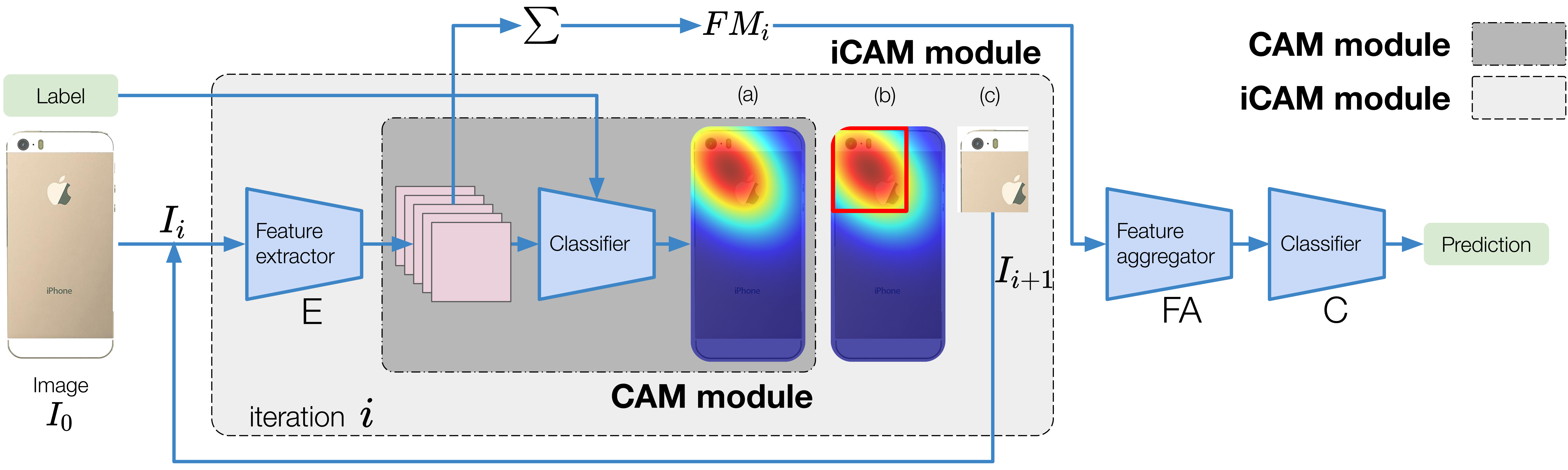}
    \end{center}
    \caption{Classification network architecture with SP1 as the input image $I_0$. For iteration $i$, an image $I_i$ is passed as an input to the $iCAM$ module. Internally, the $iCAM$ module extracts feature maps ($FM_i$) of image $I_i$ and extracts the region of activation (i.e., the region most important for classification) using the $CAM$ module. Based on the importance regions, an image $I_{i+1}$ is cropped from $I_i$, which is used for the iteration $i+1$. For the final prediction, we aggregate the feature maps in the feature aggregator ($FA$) and pass it through a classifier block to generate a prediction. Subimages (a), (b) and (c) shows the heat map, the most important region of the image given the heat map, and the cropped image from the heat map.}
    \label{fig:iCAM module}
    \vspace{-0.4cm}
\end{figure*}

\subsection{Domain Randomization}
One of the obstacles of using a machine learning algorithm is the lack of readily available, large, annotated datasets. The machine learning community has developed the concept of domain randomization (DR) to synthetically generate large datasets instead of collecting and labeling data manually. DR aims to add large variability to the synthetic data generation process, attempting to capture the real-world variations. 

Svetozar and Jianhong~\cite{valtchev2021domain} show a useful analysis of important parameters to generate synthetic data for classification for RGB images. DR has also been used in non-classification tasks. Tobin \emph{et. al.}~\cite{Tobin2017DomainRF} used DR to synthetically generate images of geometrical shapes for an autonomous robotic picking task. Additional applications include autonomous drone flight~\cite{Loquercio2019}, object detection~\cite{Tremblay2018TrainingDN}, viewpoint estimation~\cite{Movshovitz2016}, and human pose estimation~\cite{wenzheng2016}. These works demonstrate the application of DR in the RGB domain, but they do not explore their use in other domains like X-rays.  

\subsection{Fine-grained image classification}
Fine-grained image classification captures discriminative features of fine-grained objects and uses those features in image classification. In this paradigm, a localization sub-network is designed for locating the key regions of interest followed by a classification network used for recognition. Previous works on this paradigm depended on adding dense part annotations for accurate part localization and then using them for image recognition \cite{Zhang2014PartBasedRF}, \cite{Lin2015}, \cite{wei2018mask}. Recent works require only image labels for accurate part localization for training the recognition sub-network \cite{jaderberg2015spatial},\cite{fu2017look},\cite{Zheng2017MACNN}. Finally, additional techniques such as attention mechanisms \cite{yang2018learning} and multi-stage strategies \cite{he2017weakly} perform the joint training of the integrated localization-classification sub-networks. 

\subsection{Multi-domain classification}\label{AA}
Multi-domain learning aims to improve the classification performance in general domains by making full use of the information in each domain. Prior work has focused on extracting hand-crafted features such as scale-invariant feature transform~(SIFT) from the  multi-domain (RGB-D) image to get a combined image descriptor \cite{Lai2011}. Blum et al. proposed an RGB-D descriptor that relies on a k-means-based feature learning approach\cite{blum2012learned}. However, the hand-crafted features are often dataset-specific and require a strong understanding of domain-specific knowledge \cite{zheng2017sift}. To reduce the dependency on hand-crafted features, machine learning techniques are explored, and one of the most common techniques is using network parameter sharing \cite{eitel2015multimodal}.
\section{Approach}\label{Approach}
This section discusses the X-ray dataset generation process, the network architecture used for classification, and the joint training strategy used for combining the X-ray mode with the existing RGB mode for a combined RGB-X classification. In both these domains, we conduct experiments on 10 classes of modern smart phones (SP) classes.

\subsection{Datasets}
 The 10 classes of modern smart phones will be referred to as SP1, SP2, SP3, SP4, SP5, SP6, SP7, SP8, SP9, SP10 in this work. For the RGB domain, the test dataset consists of 799 real RGB images. RGB training set includes a collection of synthetic and real images. The training dataset consists of a combination of 1000 synthetic and 250 real images per class. We collect RGB images using a Microsoft Azure Kinect camera mounted over a conveyor belt. We load the mesh of the phones in Blender~\cite{blender} and use DR to create synthetic dataset by varying color, texture, and camera intrinsic and extrinsic parameters as suggested in~\cite{Tobin2017DomainRF}.

\begin{figure}[t]
    \begin{center}
        \includegraphics[width=0.6\linewidth]{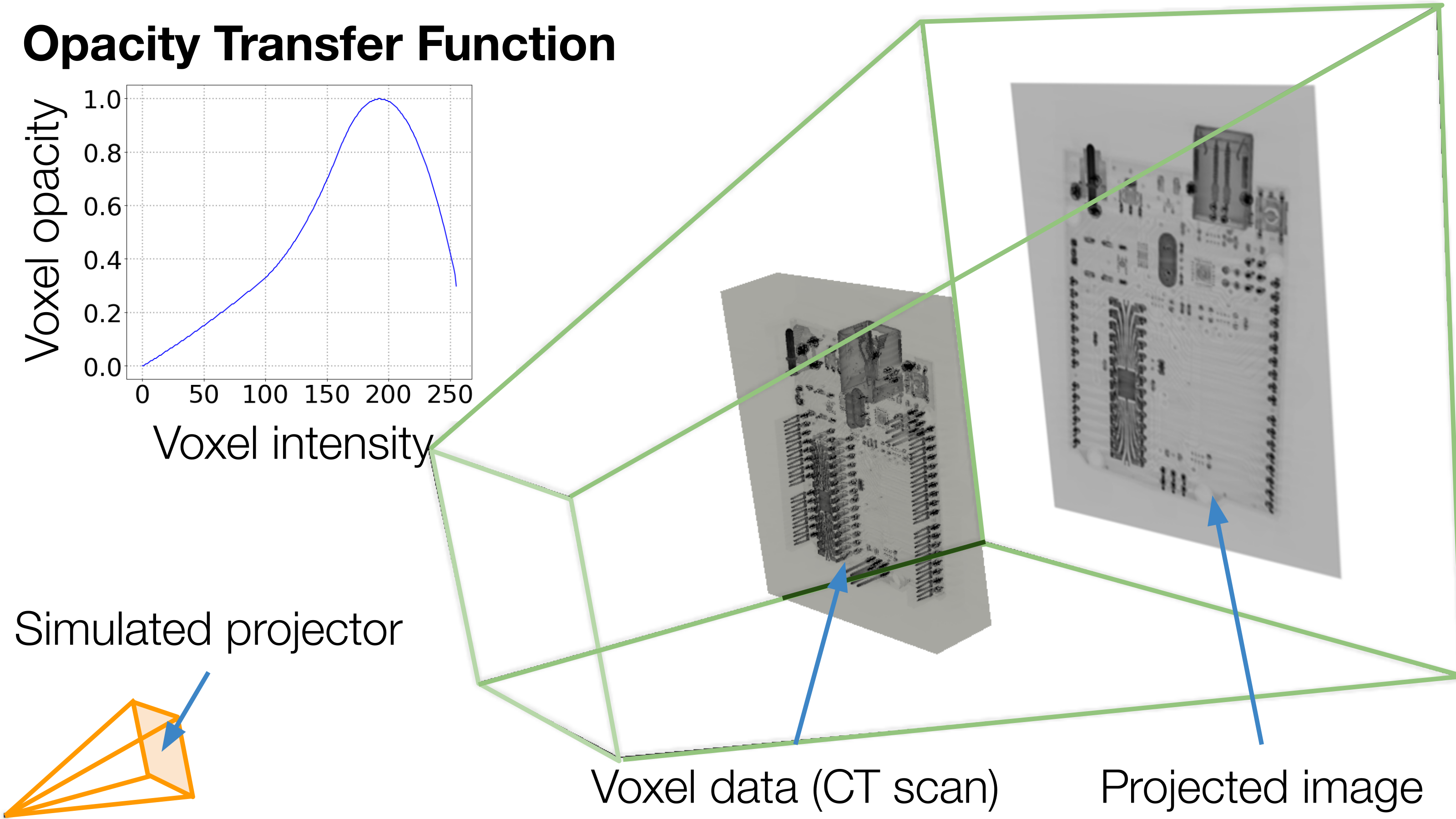}  
    \end{center}
    \caption{X-ray Domain Randomization Setup. To generate synthetic X-ray images, we project CT scans onto a plane, simulating the process of taking an X-ray image of an electronic object. We show our setup on a simple electronic device due to privacy reasons tied to the phone image.} 
    \label{fig:transf_func}
    \vspace{-0.8cm}
\end{figure}

\textcolor{black}{For the entire X-ray domain, the test dataset consists of 749 real X-ray images and around $\approx$25000 synthetic X-ray images for training. Only synthetic X-ray images are used for training to demonstrate the efficacy of domain randomization and because of the lack of enough real X-ray present for each class.} Test images are collected using an X-ray system equipped with 300kV micro focus and 180kV nano focus tubes, and a cesium iodide (CsI) detector. Both tubes are used to generate 2D and 3D images with resolutions ranging from 57.5 $\mu$m to 100 $\mu$m. The 2D and 3D images taken with the 300 kV tube use a peak energy of 220 kV with a current ranging from 200 to 300 $\mu$A, while images acquired with the 180 kV tube use 160 kV and 280 $\mu$A. For training, we utilize DR in order to generate a large scaled labelled dataset. To generate X-ray training data, we initially generate 3D voxel models of 2-3 devices per class using off the shelf CT machines. CT scans consist of a scalar intensity assigned to each voxel based on the structure of the device. The voxel intensities are mapped to an opacity value and projected onto a plane using a simulated projector model to render X-ray images. The voxel intensity to voxel opacity mapping is known as a transfer function. A variety of X-ray machine settings can be simulated by the shape of the transfer function as shown in Fig \ref{fig:transf_func}. We use VTK library~\cite{schroeder1998visualization} to randomize the transfer function, and the intrinsic and extrinsic parameters of simulated projectors.
Randomizing the transfer function arbitrarily can undesirably produce completely saturated images by assigning the same value to each pixel. 
To avoid this, we seed the randomization with 5 manually selected transfer function parameters. These manual seeds were selected using the visualization tool 3D Slicer~\cite{3DSlicer}.
We add randomness to these parameters and only select rendered images that have more than 100 SuperPoint~\cite{superpoint} keypoints. Sample training and testing images from the RGB and X-ray dataset are shown in Fig \ref{fig:dataset_image}. 
\vspace{-0.1cm}
\vspace{-0.1cm}
\subsection{Network architecture and training strategy}

This section describes the fine-grained classification network used for training the RGB and the X-ray mode, and the training strategy used in the single training and joint training phases. Let RGB image ($I_R$) or X-ray image ($I_X$) be  of dimension $\real{H}\times \real{W}\times \real{C}$ in the labeled dataset available for training the classification model, and $y_i$ be the image label in one-hot encoding - \emph{i.e.}, $y_i \in  R^m$ is a vector of dimensionality \textit{m} (the number of classes) with \textcolor{black}{$y_{i}^{j=l}$ = 1 and $y_{i}^{j \neq l}$ = 0 for the position \textit{l}} denoting the image label. We train the model using a three-stage approach: first training the 2 stream networks individually, followed by a joint fine-tuning stage. 

\subsubsection{Iterative class activation mapping (iCAM)}
As shown in Fig \ref{fig:iCAM module}, our network has 4 modules: feature extractor module ($E$), class activation mapping module ($CAM$), feature aggregator module ($FA$), and classifier module ($C$). The feature extractor module ($E$) and the $CAM$ module are stacked together in an iterative manner followed by the feature aggregator module and classifier module.

In our method, we use class activation mapping (CAM)~\cite{zhou2015cnnlocalization} using the global average pooling (GAP) to indicate the discriminative image regions used by CNNs to identify the respective classes. For a given image $I \in \mathbb{R}^{H\times W\times C}$, let $FM(x,y) = \sum_{j=1}^kfm_j(x, y)$ represent the activation of the last convolutional layer at spatial location $(x,y)$ with $k$ channels. $\theta_C^c$ denotes the weight corresponding to class c in the classifier module C. We define $m_c$ as the class activation map for class c as:
\begin{equation}
    \begin{split}
        &fm_{0\dots k}(x,y) = f_{\theta_E}(I), \\
        &m_c(x,y)  = \sum_{j=1}^k{\theta_C^cfm_i(x,y)}.
        \label{cam}
    \end{split}
\end{equation}
Thus, $m_c(x, y)$ is an unnormalized probability distribution directly indicating the importance of the activation at spatial location $(x, y)$. More details on CAM can be found in \cite{zhou2015cnnlocalization}.

In our experiments, InceptionNet-v3~\cite{szegedy2016rethinking} up to the Mixed7c layer is chosen as the feature extraction module with $\theta_E$ as a learnable parameter. The feature map ($FM_i$) and the class activation map ($m_{c,i}$), for every iteration $i$, is calculated using Eq.\ref{cam}. 
The activation map is upsampled to the original image size $H\times W\times C$, and then a square region around the most dominant pixel of the map is selected resulting in image $I_{i+1}$. The cropped image is then upsampled to the original image dimension $H\times W\times C$ using bilinear upsampling function in Pytorch.
After running the feature extraction and the CAM module for a pre-defined number of iterations $n$, $F_{0..n}$ is passed to the feature aggregator module followed by the classifier module to get the final classification score $S$:
\vspace{-0.1cm}
 \begin{equation}
    \begin{split}
               &FM(x,y)  = \sum_{i=1}^n{\theta_{FA}FM_{i}(x,y)}, \\
               &S =  \mbox{softmax}(f_{\theta_{C}}(FM)),
    \end{split}
\end{equation} where $FM$ is the aggregated feature map, and $\theta_{FA}$ and $\theta_C$ are the learnable parameters of the feature aggregator and the classifier module. For our experiments, we choose n=3.  \\
\vspace{-0.15cm}
\subsubsection{Training strategy}
\label{joint_training}
We proceed by training the iCAM network to minimize the negative log likelihood, $\mathcal{L}$, of the training data. While training we solve for:

\vspace{-0.5cm}
\begin{multline}
        \theta_{E}^*,\theta_{FA}^*,\theta_{C}^* = \\ \mbox{argmin}_{\theta_{E},\theta_{FA},\theta_{C}}\sum_{i=1}^{N}\mathcal{L}(f_{iCAM}(I,y_i;\theta_{E},\theta_{FA},\theta_{C}),y_i),
\end{multline}

We individually train the $iCAM_R$ and $iCAM_X$ for $I_{R}$ and $I_{X}$ images on RGB and X-ray datasets. For both the streams, the $\theta_E$ is initialized with InceptionNet-v3 weight pretrained on ImageNet and were trained for 40 epochs each. The implementation is highly parallelized and performs full-batch gradient descent using the Stochastic Gradient Descent~\cite{bottou2010large} optimizer in the Pytorch Autograd library ~\cite{paszke2017automatic}, with a batch size of 16 with a learning rate of 0.001. At the end of this individual training stage, $iCAM_R$ and $iCAM_X$ have different sets of weights which are later fused in the joint training stage.

Once both domains are trained individually, we use the parameter sharing method to combine the information of both the domains as shown in Fig\ref{fig:pipeline_image}. During the joint-training, the softmax activation is discarded and the output from the classification layers are concatenated. Their individual responses $FM_{R}$ from $iCAM_R$, and $FM_{X}$ from $iCAM_X$, are fused and fed through an additional stream $FS([g_R,g_X])$ with parameters $\theta_{FS}$. This fusion network again ends in a
softmax classification layer. During the joint-training phase, $\theta_{E,R}$, $\theta_{FA,R}$, $\theta_{E,X}$ and $\theta_{FA,X}$ are frozen and only $\theta_{C,R}$, $\theta_{C,X}$ and $\theta_F$ are open to training. In this stage of training, we solve:
\vspace{-0.3cm}
\begin{multline}
\theta_{C,R}^*,\theta_{C,X}^*,\theta_{FS}^*=
\\
\mbox{argmin}_{\theta_{C,R},\theta_{C,X},\theta_{FS}}\sum_{i=1}^{N}\mathcal{L}(f_{FS}([g_R(I_{R}),g_X(I_{X})],y_i).        
\end{multline}

During the joint-training phase, only the weights of the fusion layer and the weights of individual classification layers are optimized, keeping all other weights from individual mode intact.
\section{Analysis and Results} \label{results}

\subsection{Domain randomization}

In this set of experiments, we evaluate the effect of domain randomization parameters on test accuracy. 
Our current implementation randomizes image white noise, background color, transfer function parameters, and rendering projector position and orientation. 
As shown in Table~\ref{tab:Fixed_DR_params} in the first experiment, we introduce noise of increasing variance in the rendered image.
\textcolor{black}{We observe that the best accuracy of $98.1\%$ is obtained when $\sigma=1000$ with varying background, transfer function, and projector position and orientation. Also, we see a drop in the accuracy with the drop in the noise variance. The presence of noise improves convergence and makes
training less susceptible to local minima as shown in \cite{DBLP:journals/corr/TobinFRSZA17}}. We also observe that fixing the projector position and orientation affects accuracy the most, lowering it down to only $31.5\%$.
This result emphasizes the use of DR as this projector position simulates different views of a 3D object, which can not be simulated by simple 2D data augmentation. 
Another important factor is the transfer function, which as discussed in Sec.~\ref{Approach}, simulates different X-ray machines and X-ray machine settings. 
Varying the transfer function, generalizes our data over different machine distributions. Unlike the RGB domain, we avoid pattern randomization since such variations do not exist in our test data.
\subsection{iCAM}
\begin{table}[]
\begin{center}
\caption{Effect of DR parameters on accuracy. \textcolor{black}{Current DR has non-fixed background, non-fixed transfer function, non-fixed simulated projector position and orientation and $\sigma=1000$ as the variance for the white noise.}}
\label{tab:Fixed_DR_params}
\begin{adjustbox}{width=0.85\linewidth}
\begin{tabular}{|c|c|c|}
\hline
\multicolumn{1}{|c|}{\textbf{Parameters}} & \multicolumn{1}{c|}{\textbf{Status}} & \multicolumn{1}{c|}{\textbf{Accuracy}} \\ \hline
Current DR & Sec.~\ref{Approach} &0.981  \\ \hline
\multirow{3}{*}{Noise} & $\sigma$=10 & 0.934  \\ \cline{2-3} 
 & $\sigma$=100 & 0.923 \\ \cline{2-3} 
 & $\sigma$=500 & 0.957 \\ \hline
Background & Fixed & 0.922 \\ \hline
\begin{tabular}[c]{@{}c@{}}Transfer  function\end{tabular} & Fixed & 0.655 \\ \hline
\begin{tabular}[c]{@{}c@{}}Simulated projector position\\ and orientation\end{tabular} & Fixed & 0.315  \\ \hline
\end{tabular}
\end{adjustbox}
\end{center}
\vspace{-0.95cm}
\end{table}

\textcolor{black}{iCAM is a fine grained classification method. For the sake of result section, we show the performance of iCAM on the modern smart phones (SP) classes mentioned in Section III and also CUB-200\cite{wah2011caltech} dataset.} 
For comparisons, we evaluate the results with baseline classification networks like VGG-19\cite{simonyan2014very}, ResNet-101 \cite{he2016deep}, ResNeXt-101\cite{xie2017aggregated} and InceptionNet-v3\cite{szegedy2015rethinking}. We also compare it with the SqueezeExcitation\cite{hu2018squeeze} with ResNet101 as the base module. All these experiments are performed on the X-ray dataset. We use average accuracy, and inference time as the metrics to compare the results for different networks as suggested in \cite{Bianco_2018}. We also compare average precision and average recall of all these network architecture to our method as suggested in \cite{grandini2020metrics}. These results are presented in Table \ref{network-arch}. 
Additionally, we compare the iCAM+InceptionNet-v3 to these baseline classification networks on the CUB-200 dataset \cite{wah2011caltech}, a standard fine-grained classification dataset used for bench-marking image classification algorithms. 

Table \ref{network-arch} shows that the iCAM block when combined with the InceptionNet-v3 base module outperforms the baseline image classification networks. The high accuracy value is attributed to the hierarchical approach iCAM takes in order to extract and combine key features from an image. Intuitively this approach is similar to looking and extracting key features at different levels of magnification and then using  all the features for classifying that image. For further analysis we calculate the heat-map using CAM as seen in Fig \ref{fig:heat_map}, and use them to visualize the CNN's area of interest for similar looking classes (e.g., SP2 vs SP3). It is seen that iCAM block indeed helps in localizing key features in an X-ray image. Additionally, Table \ref{network-arch} presents that iCAM+InceptionNet-v3 has an inference time of $\approx$30fps, which allows the model to run in real-time with a 30fps industrial camera. Table~\ref{CUB datset}, shows that when iCAM is combined with InceptionNet-v3, the network outperforms InceptionNet-v3 by 10.5$\%$. We also combine the iCAM block with the ResNet-101 feature extractor, and the result outperforms ResNet-101\cite{he2016deep}, and other networks like ResNeXt-101\cite{xie2017aggregated}, SE-ResNet-101\cite{hu2018squeeze}, and VGG-19\cite{simonyan2014very}. This shows the efficacy of iCAM block with other extraction modules.
\vspace{1cm}
\begin{table}[]
\begin{center}
\caption{Comparison of baseline classification networks to the iCAM network on the X-ray dataset. Learning rate: 0.001, Batch Size: 16, Weight decay: 0.1 every 7 epochs, Epochs: 40. \textcolor{black}{The bolded row highlights the network architecture with the best average accuracy.}}
\label{network-arch}
\begin{adjustbox}{width=0.85\linewidth}
\begin{tabular}{|@{\hskip3pt}c@{\hskip3pt}|@{\hskip3pt}c@{\hskip3pt}|@{\hskip3pt}c@{\hskip3pt}|@{\hskip3pt}c@{\hskip3pt}|@{\hskip3pt}c@{\hskip3pt}|}
\hline
\textbf{\begin{tabular}[c]{@{}c@{}}Network\\Architecture\end{tabular} } & \textbf{\begin{tabular}[c]{@{}c@{}}Average\\accuracy\end{tabular}} & \textbf{Precision} & \textbf{Recall} & \textbf{\begin{tabular}[c]{@{}c@{}}Infer\\time (ms)\end{tabular} }\\ \hline

InceptionNet-v3        & 0.933                & 0.907                 &   0.935       & 12.3       \\ \hline
ResNet-101           & 0.953                & 0.899                  & 0.956        &11.1     \\ \hline
VGG-19              & 0.956               & 0.898                    & 0.961        &7.53   \\ \hline
\begin{tabular}[c]{@{}c@{}}ResNeXt-101\\  (64x4d)\end{tabular}           & 0.929               & 0.869                    & 0.948       &20.15     \\ \hline
SE-ResNet-101             & 0.891               & 0.841                  & 0.926         & 14.01   \\ \hline
\textbf{\begin{tabular}[c]{@{}c@{}}iCAM+\\  InceptionNet-v3\end{tabular}}     & \textbf{0.979}        & \textbf{0.942}                 & \textbf{0.969}   & \textbf{30.86}          \\ \hline

\end{tabular}
\end{adjustbox}
\end{center}
\vspace{-0.4cm}
\end{table}

\begin{table}[]

\begin{center}
\caption{Comparison of baseline classification networks to the iCAM network on the CUB-200\cite{wah2011caltech} dataset. Same parameters as Table 2. \textcolor{black}{The bolded row highlights the network architecture with the best average accuracy for InceptionNet-v3 and ResNet-101 feature extractor.}}
\label{CUB datset}
\begin{adjustbox}{width=0.8\linewidth}
\begin{tabular}{|c|c|c|c|c|c|c|}
\hline
\textbf{\begin{tabular}[c]{@{}c@{}}Network\\Architecture\end{tabular}} & \textbf{Accuracy} & \textbf{Precision} & \textbf{Recall}\\ \hline

InceptionNet-v3        &0.692&0.697&0.693\\ 
\textbf{\begin{tabular}[c]{@{}c@{}}iCAM+\\  InceptionNet-v3\end{tabular}}     &\textbf{0.797}&\textbf{0.791}&\textbf{0.801}\\ \hline
ResNet-101           &0.814&0.818&0.815\\ 
\begin{tabular}[c]{@{}c@{}}ResNeXt-101\\  (64x4d)\end{tabular}           &0.829&0.835&0.830\\ 
SE-ResNet-101             &0.818&0.826&0.820\\ 
\textbf{\begin{tabular}[c]{@{}c@{}}iCAM+\\  ResNet-101\end{tabular}}     &\textbf{0.834} &\textbf{0.831} &\textbf{0.838} \\ 
VGG-19              &0.778&0.781&0.779\\ \hline

\end{tabular}
\end{adjustbox}
\end{center}
\vspace{-0.6cm}
\end{table}

\begin{figure}[t]
    \begin{center}
        \includegraphics[width=\linewidth]{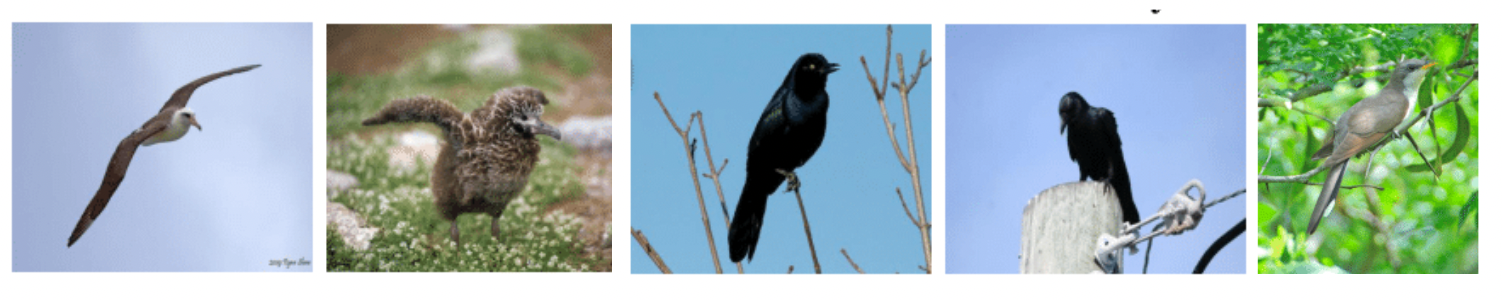}  
    \end{center}
    \caption{\textcolor{black}{Examples of training and testing images from the CUB-200\cite{wah2011caltech} dataset. Small inter-class variance and large intra-class variance between species motivates us to use this dataset for benchmarking iCAM.}} 
    \label{fig:heat_map}
    \vspace{-0.1cm}
\end{figure}

\begin{figure}[t]
    \begin{center}
        \includegraphics[width=0.95\linewidth]{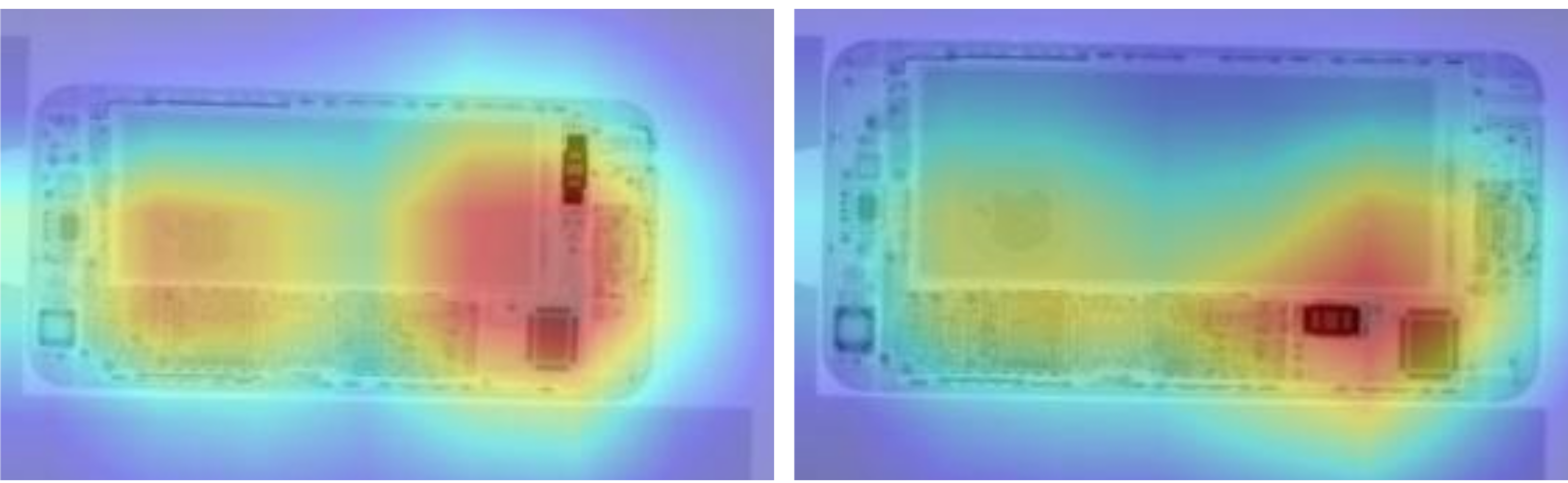}  
    \end{center}
    \caption{Heatmap for SP2 (left) and SP3 (right) generated using CAM. SP2 has a prominent logo visible in the X-ray image, that is the dominant region for classification. SP3 has a logo and a different component, that helps identify SP3.} 
    \label{fig:heat_map}
    \vspace{-0.3cm}
\end{figure}

We also conduct a study on the number of iterations the iCAM module is run during the training time. Table \ref{iCAM N iter} shows the accuracy, the time taken to train 40 epochs, and the inference time for $n=2,3,5$. As shown in the table, $n=3$ shows a considerable increase in accuracy from using $n=2$ for far less training and inference time that $n=5$.

\begin{table}[]
\begin{center}
\vspace{-0.3cm}
\caption{Comparison of accuracy, training time and inference time of the iCAM+InceptionNet-v3 network for different values of iteration N. Training was done for 40 epochs.\textcolor{black}{ The bolded row highlights the current $n$ chosen for all the experiments.}}
\label{iCAM N iter}
\begin{adjustbox}{width=\linewidth}
\begin{tabular}{|@{\hskip3pt}c@{\hskip3pt}|c|c|c|c|}

\hline
\textbf{\begin{tabular}[c]{@{}c@{}}No. of CAM\\  iteration (N)\end{tabular}} & \multicolumn{1}{c|}{\textbf{\begin{tabular}[c]{@{}c@{}}Average\\ accuracy\end{tabular}}}& \multicolumn{1}{c|}{\textbf{\begin{tabular}[c]{@{}c@{}}Training\\ time (mins)\end{tabular}}} & \multicolumn{1}{c|}{\textbf{Infer-time(ms)}} \\ \hline
n=2                              & 0.962                  & 325                                    &  20.11              \\ \hline
\textbf{n=3}                              & \textbf{0.979}        & \textbf{490}                 & \textbf{30.86}                \\ \hline
n=5                              &  0.981                 & 723                                    &  50.96              \\ \hline
\end{tabular}
\end{adjustbox}
\end{center}
\vspace{-0.7cm}
\end{table}

A comparative study is done to select the best feature aggregator strategy where 3 different feature aggregator strategies are tested. The most common strategies used are mean, sum, and weighted mean of the available feature maps. All of these experiments are done for $n=3$. We achieve an accuracy of $0.941$ for the mean, $0.768$ for the sum, and $0.979$ for the weighted mean version of the feature aggregator. Based on this study, we conclude that weighted mean is the best feature aggregating strategy because it weighs different level of magnification differently . 

\vspace{-0.1cm}
\subsection{RGB-X classification}
We evaluate the multi-modal classification approach with both the single-mode iCAM+InceptionNet-v3 architectures. For this experiment, an RGB image of a particular class is paired with a random X-ray image sampled from the same class. This is done because of the lack of a streamlined RGB + X-ray sensor setup. While pairing the images, we randomly re-orient the X-ray images to prevent any bias induced due to similar orientation of the RGB and X-ray image. \textcolor{black}{The total training time for RGB-X mode is $520$ mins as the RGB and X-ray modes are trained in parallel.} Table \ref{multimodal accuracy} shows the average accuracy, precision, recall \textcolor{black}{and the inference time} of each of the mode (RGB and X-ray) as well as the combined RGB-X mode.
\textcolor{black}{
In context of a recycling plant that recycles around 200 phones per hour \cite{Daisy2019},
the 0.6 percentage gain in accuracy from RGB-X ($n=3$) over X-ray ($n=5$) is critical as it saves 12 phones per hour. Although this adds $~11ms$ to the inference time, the total training time is lowered to $520$ mins (RGB-X) from $723$ mins (X-ray).
The lower training time is crucial while expanding to newer products without compromising on the recycling rate of 200 phones per hour.}



\begin{table}[]
\begin{center}
\caption{Comparison of single (RGB, X-ray) mode accuracy to the multi-modal (RGB-X mode) accuracy. \textcolor{black}{The bolded row highlights the mode with the best average accuracy.}}
\label{multimodal accuracy}
\begin{adjustbox}{width=1.0\linewidth}
\begin{tabular}{|c|c|c|c|c|}
\hline
\textbf{Mode}  & \multicolumn{1}{c|}{\textbf{\begin{tabular}[c]{@{}c@{}}Average\\ accuracy\end{tabular}}} & \multicolumn{1}{c|}{\textbf{\begin{tabular}[c]{@{}c@{}}Average\\ precision\end{tabular}}} & \multicolumn{1}{c|}{\textbf{\begin{tabular}[c]{@{}c@{}}Average \\ recall\end{tabular}}} &
\multicolumn{1}{c|}{\textbf{\begin{tabular}[c]{@{}c@{}}Infer\\time (ms)\end{tabular}}} \\ \hline
RGB            & 0.891                                                                                         &  0.932                                                                                        & 0.863
&30.86\\ \hline
X-ray          & 0.979                                                                                  & 0.942                                                                                         &  0.969 
&30.86\\ \hline
\textbf{RGB-X} & \textbf{0.987}                                                                                         & \textbf{0.987}                                                                                          & \textbf{0.995}    
        & \textbf{62.4}\\ \hline
\end{tabular}
\end{adjustbox}
\end{center}
\vspace{-0.4cm}
\end{table}
As mentioned in Sec.\ref{joint_training}, for joint training we combine the deepest layers and train for them, by keeping the higher CNN layers frozen. This fusion approach is termed as ``fast fusion'' \cite{shaikh2021rgb}. We evaluate this approach with the ``slow fusion'' approach also mentioned in \cite{shaikh2021rgb}. For slow fusion, the network architecture is similar to fast fusion, except for the fact that the weights of the higher CNN layers are also trainable. This allows the higher layers to have access to more global information.

\begin{table}[]
\begin{center}
\caption{Comparison of different fusion techniques. \textcolor{black}{The bolded row highlights the mode with the best average accuracy.}}
\label{fusion technique accuracy}
\begin{adjustbox}{width=0.86\linewidth}
\begin{tabular}{|c|c|c|c|}
\hline
\textbf{Mode}        & \multicolumn{1}{c|}{\textbf{\begin{tabular}[c]{@{}c@{}}Average\\ accuracy\end{tabular}}} & \multicolumn{1}{c|}{\textbf{\begin{tabular}[c]{@{}c@{}}Average\\ precison\end{tabular}}} & \multicolumn{1}{c|}{\textbf{\begin{tabular}[c]{@{}c@{}}Average \\ recall\end{tabular}}} \\ \hline

Slow Fusion & 0.952 & 0.951 & 0.958 \\ \hline
\textbf{Fast Fusion} & \textbf{0.987} & \textbf{0.987} & \textbf{0.995} \\ 
\hline 
\end{tabular}
\end{adjustbox}
\end{center}
\vspace{-0.5cm}
\end{table}

Table \ref{fusion technique accuracy}, shows the comparisons of 2 different fusion techniques, demonstrating that fast fusion performs better than the slow fusion, as fast fusion doesn't affect the lower level activation learned by the iCAM layers.
\vspace{-0.1cm}
\section{Conclusion and Future work} \label{conclusion}
We introduce a useful multi-modal neural network architecture for the RGB-X object for the task of WEEE classification. Our method consists of a two-stream convolutional neural network that learns to fuse information from both RGB and X-ray domains automatically before classification. We also present a novel fine-grained network architecture, which is used to train the individual streams of data. This method iteratively localizes key features in an object and uses them for classification. Our experiments explore how the various components of this algorithm achieve better classification than the baseline algorithms when tested on 10 classes of modern smart phones. \textcolor{black}{For this paper, we use iPhone as our modern smart phones, as the devices and the dataset were readily available.} We also present a novel domain randomization pipeline for X-ray images using CT scans, a method that reduces the burden of annotating thousands of X-ray images. 
\textcolor{black}{As our future work, we plan to extend the work to classify smart phones from different manufacturers like Samsung, Motorola and also between inter-manufacturer models.}
Additionally, we plan to extend the approach to attend to multiple important regions in an activation map. We also plan to generalize the hierarchical attention approach to other interpretable methods like Grad-CAM\cite{selvaraju2017grad}.

The high accuracy as well as high inference rate of this approach make it well suited for integration into electronics sorting processes at material recovery facilities.



\bibliographystyle{IEEEtran}
\bibliography{references}

\end{document}